\documentclass[10pt]{article}

\renewcommand{\paragraph}[1]{\par\vspace*{.5\baselineskip}\noindent\textbf{#1}}

\usepackage{iwpt97}
\usepackage{epsfig}
\usepackage{nobibib}

\newcommand{\meth}[1]{{\small \sffamily  #1}}

\newcommand{\bsp}[1]{\textsl{``#1''}}

\title{Message-Passing Protocols for Real-World Parsing ---\\
An Object-Oriented Model and its Preliminary Evaluation}

\author{Udo Hahn and Peter Neuhaus and Norbert Br{\"o}ker
\\
{\small \raisebox{-0.25mm}{\epsfxsize=4ex\epsfbox{clif-logo.eps}}}
Computational Linguistics Lab \\
Freiburg University,
Werthmannplatz,
D-79085 Freiburg, Germany\\
\texttt{\small\{hahn,neuhaus,nobi\}@coling.uni-freiburg.de}\\
\texttt{\small http://www.coling.uni-freiburg.de}}

\date{}

\begin{document}

\maketitle


\vspace*{6mm} 
\begin{abstract}
We argue for a performance-based design of natural language grammars
and their associated parsers in order to meet the constraints imposed
by real-world NLP. Our approach
incorporates declarative and procedural knowledge about language and
language use within an object-oriented specification framework. We
discuss several message-passing protocols for parsing and provide
reasons for sacrificing completeness of the parse in favor of
efficiency based on a preliminary empirical evaluation.
\end{abstract}

\vspace*{3mm}



\section{Introduction}\label{ch:intro}

Over the past decades the design of natural language grammars and
their parsers was almost entirely based on {\em competence}
considerations \cite{chomsky65}.  These hailed pure declarativism
\cite{shieber86} and banned procedural aspects of natural language use
out of the domain of language theory proper.  The major premises of
that approach were to consider sentences as the primary 
object of linguistic investigation, to focus on syntactic descriptions, and to
rely upon perfectly well-formed utterances for which complete grammar
specifications of arbitrary depth and sophistication were available.
In fact, promising efficiency results can be achieved for parsers operating 
under corresponding optimal laboratory conditions.  
Considering, however, the requirements of natural language
{\em understanding}, i.e., the integration of syntax, semantics, and
pragmatics, and taking {\em ill-formed} input or {\em incomplete}
knowledge into consideration, their processing costs either tend to
increase at excessive 
rates or linguistic processing even fails completely.

As a consequence, the challenge to meet the specific requirements
imposed by real-world texts has led many researchers in the NLP
community to re-engineer competence grammars and their parsers and to
provide various add-ons in terms of constraints \cite{uszkoreit91},
heuristics \cite{huyck93}, statistics-based weights \cite{charniak93},
etc.  In contradistinction to these approaches, our principal goal has been 
to incorporate performance conditions already in
the design of natural language grammars, yielding so-called {\em
performance grammars}. 
%
%
Thus, not only declarative knowledge (as is common for competence
grammars), but also {\em procedural} knowledge (about control and
parsing strategies, resource limitations, etc.) has to be taken into
consideration at the {\em grammar specification} level proper. This is
achieved by providing self-contained description primitives for the
expression of procedural knowledge. We have taken care to
transparently separate declarative (structure-oriented) from
procedural (process-oriented) knowledge pieces. 
Hence, we have chosen a formally
homogeneous, highly modularized object-oriented grammar specification
framework, {\em viz.} the actor model of computation which is based on
concurrently active objects that communicate by asynchronous message
passing \cite{agha90}. 

The 
parser whose design is based on these performance
considerations 
forms part of a text knowledge
acquisition system,
operational in two domains, {\em viz.}
the processing of test reports from the information technology field \cite{hahn.kdd97}
and medical reports \cite{hahn.amia96}.  
The analysis of texts (instead of isolated
sentences) requires, first of all, the consideration of textual
phenomena by a dedicated {\em text grammar}. 
Second, text understanding
is based on drawing {\ inferences} by which text propositions
are integrated on the fly into the text knowledge base with reference to a
canonical representation of the underlying {\em domain knowledge}. 
This way, grammatical (language-specific) and
conceptual (domain-specific) knowledge are closely coupled.  Third,
text understanding in humans occurs immediately and at least within
specific processing cycles in parallel \cite{thibadeau82}.
These processing strategies we find in human language processing are
taken as hints how the complexity of natural language
understanding can reasonably be overcome by machines.
Thus, text parsing devices should operate {\em incrementally}
and  {\em concurrently}.
In addition, the consideration of {\em real-world} texts 
forces us to supply mechanisms which allow for the {\em robust} processing
of extra- and ungrammatical input.
We take an approach where ---
in the light of abundant specification gaps at the grammar and 
domain representation level ---
the degree of underspecification of the knowledge sources or the impact
of grammar violations directly corresponds to a lessening of the precision
and depth of text knowledge representations, thus aiming at a sophisticated
{\em fail-soft} model of {\em partial} text parsing.


\section{The Grammar}
\label{grammar}

The 
performance grammar we consider contains fully {\em lexicalized}
grammar specifications \cite{hahn940}. Each lexical item is subject to
configurational constraints on word classes and morphological features
as well as conditions on word order and conceptual compatibility a
head places on possible modifiers.
Grammatical conditions of these types are combined in terms of {\em
valency} constraints (at the phrasal and clausal level) as well as
{\em textuality} constraints (at the text level of consideration),
which concrete dependency structures and local as well as global
coherence relations must satisfy. The compatibility of grammatical
features including order constraints (encapsulated by methods we refer
to as \textsc{syntaxCheck}) is computed by a unification mechanism,
while the evaluation of semantic and conceptual constraints (we here
refer to as \textsc{conceptCheck}) relies upon the terminological and
rule-based construction of a consistent conceptual representation.
Thus, while the dependency
relations represent the linguistic structure of the input, the
conceptual relations yield the targeted  representation of the text
content (for an illustration, cf. Fig.\ \ref{kb}).

In order to structure the underlying lexicon, {\em inheritance}
mechanisms are used. Lexical specifications are organized along
the grammar hierarchy at various abstraction levels, e.g., with
respect to generalizations on word classes. Lexicalization of this
form already yields a fine-grained decomposition of declarative
grammar knowledge. It lacks, however, an  equivalent description
at the procedural level.  We therefore provide lexicalized
communication primitives to allow for heterogeneous and local forms of
interaction among lexical items.

Following the arguments brought forward, e.g., by
\textcite{jackendoff90} and \textcite{allen93}, there is no
distinction at the representational level between semantic and
conceptual interpretations of texts. Hence, semantic and domain
knowledge specifications are based on a common hybrid
classification-based knowledge representation language (for a survey,
cf.\ \textcite{woods+schmolze92}).  Ambiguities which result in
interpretation variants are managed by a context mechanism of the
underlying knowledge base system.

{\em Robustness} at the grammar level is achieved by several
means. Dependency grammars describe binary, functional relations
between words rather than contiguous constituent
structures. Thus, ill-formed input often has an
(incomplete) analysis in our grammar. Furthermore, it is possible to
specify lexical items at different levels of syntactic or semantic
granularity such that the specificity of constraints may vary.
The main burden of robustness, however, is
assigned  to a dedicated message-passing
protocol we will discuss in the next section.



\section{The Parser}


Viewed from a parsing perspective, we represent lexical items as
\emph{word actors} which are acquainted with other actors representing
the heads or modifiers in the current utterance.  A specialized actor
type, the \emph{phrase actor}, groups word actors which are connected
by dependency relations and encapsulates administrative information
about each phrase.
A message does not have to be sent directly to a specific word actor,
but will be sent to the mediating phrase actor which forwards it to an
appropriate word actor.
Furthermore, the phrase actor holds the communication channel to the
corresponding interpretation context in the domain knowledge base
system.
A \emph{container actor} encapsulates several phrase actors that constitute
alternative analyses for the \emph{same} part of the input text (i.e.,
structural ambiguities). Container actors play a central role in
controlling the parsing process, because they keep information about the
\emph{textually} related (\emph{preceding}) container actors holding the
left context and the \emph{chronologically} related (\emph{previous})
container actors holding a part of the head-oriented parse history.




\paragraph{Basic Parsing Protocol (incl.\ Ambiguity Handling).}
\label{pc:basic}
We use a graphical description language to sketch the message-passing
protocol for establishing dependency relations as depicted in
Fig.~\ref{ambig} (the phrase actor's active head is visualized by
$\bigoplus$).  A \meth{searchHeadFor} message (and {\em vice versa} a
\meth{searchModifierFor} message if \meth{searchHeadFor} fails) is
sent to the textually preceding container actor (precedence relations
are depicted by bold dashed lines), which simultaneously directs this
message to its encapsulated phrase actors. At the level of a single
phrase actor, the distribution of the \meth{searchHeadFor} message
occurs for all word actors at the ``right rim'' of the dependency tree
(depicted by \raisebox{-0.5ex}{\epsfig{file=blob1.eps,width=1em}}%
).  A word actor that receives a \meth{searchHeadFor} message from
another word actor concurrently forwards this message to its head (if
any) and tests in its local site whether a dependency relation can be
established by checking its corresponding valency constraints
(applying \textsc{syntaxCheck} and \textsc{conceptCheck}).  In case of
success, a \meth{headFound} message is returned, the sender and the
receiver are copied (to enable alternative attachments in the
concurrent system, i.e., no destructive operations are carried out),
and a dependency relation, indicated by a dotted line, is established
between those copies which join into a phrasal relationship
(for a more detailed description of the underlying protocols, cf.\
 \textcite{neuhaus.coling96}).  For
illustration purposes, consider the analysis of a phrase like
\bsp{Zenon sells this printer} covering the content of the phrase
actor which textually precedes the phrase actor holding the dependency
structure for \bsp{for \$2,000}. The latter actor requests its
attachment as a modifier of some head.  The resultant new container
actor (encapsulating the dependency analysis for \bsp{Zenon sells this
printer for \$2,000} in two phrase actors) is, at the same time, the
historical successor of the phrase actor covering the analysis for
\bsp{Zenon sells this printer}.

The structural ambiguity inherent in the example is easily accounted
for by this scheme.  The criterion for a structural ambiguity to
emerge is the reception of at least two positive replies to a single
\meth{searchHeadFor} (or \meth{searchModifierFor}) message by the
initiator. The basic protocol already provides for the concurrent
copying and feature updates.
In the example from Fig.~\ref{ambig}, two alternative readings are
parsed, one phrase actor holding the attachment to the verb
(\bsp{sells}), the other holding that to the noun (\bsp{printer}).
The crucial point about these ambiguous syntactic structures is that
they have conceptually different representations in the domain
knowledge base. In the case of Fig.\ \ref{ambig} verb attachment leads
to the instantiation of the {\sc price} slot of the corresponding {\sc
Sell} action, while the noun attachment leads to the corresponding
instantiation of the {\sc price} slot of {\sc Printer}.

\begin{figure}
\centering
\epsfig{file=ambig.eps,width=100mm}
\caption{Basic Mode (incl. Structural Ambiguities)
	\label{ambig}}
\end{figure}

\paragraph{Robustness: Skipping Protocol.}
\label{pc:skip}
Skipping for robustnes purposes is a well known mechanism though
limited in its reach \cite{lavie+tomita93}. But in free word-order
languages as German skipping is even vital for the analysis of
entirely well-formed structures, e.g., those involving
scrambling or discontinuous constructions. 
For brevity, we will base the following
explanation on the robustness issue and refer the interested reader to
\textcite{neuhaus+broker97}. 
The incompleteness of linguistic and conceptual specifications is
ubiquitous in real-world applications and, therefore, requires
mechanisms for a fail-soft parsing behavior.
Fig.~\ref{skipping} illustrates a typical ``skipping''
scenario.  The currently active container addresses a
\meth{searchHeadFor} (or \meth{searchModifierFor}) message to its
textually immediately preceding container actor.  If {\em both} types
of messages fail, the immediately preceding container of the active
container forwards these messages --- in the canonical order --- to
its immediately preceding container actor.  If any of these two
message types succeeds after that mediation, a corresponding
(discontinuous) dependency structure is built up.  Furthermore, the
skipped container is moved to the left of the newly built container
actor. Note that this behavior results in the reordering of the
lexical items analyzed so far such that
skipped containers are continuously moved to the left.  As an example,
consider the phrase \bsp{Zenon sells this printer} and let us further
assume \bsp{totally} to be a grammatically unknown item which is
followed by the occurrence of \bsp{over-priced} as the active
container. Skipping yields a structural analysis for \bsp{Zenon sells
this printer over-priced}, while \bsp{totally} is simply discarded
from further consideration.  This mode requires an extension of the
basic protocol in that \meth{searchHeadFor} and
\meth{searchModifierFor} messages are forwarded across non-contiguous
parts of the analysis when these messages do not yield a positive
result for the requesting actor relative to the {\em immediately}
adjacent container actor.

\begin{figure}
\centering
\epsfig{file=skip.eps,width=100mm}
\caption{Skipping Mode 
	\label{skipping}}
\end{figure}

\paragraph{Backtracking Protocol.}
\label{pc:back}
Backtracking to which we still adhere in our model of
constrained concurrency accounts for a state of the analysis
where none of the aforementioned protocols have terminated successfully in
{\it any textually} preceding container, i.e., several repeated
skippings have occurred, until a linguistically plausible barrier is
encountered.
In this case, backtracking takes place and messages are now directed
to {\it historically} previous containers, i.e., to containers holding
fragments of the parse history. 
This is realized in terms of a protocol extension by which
\meth{searchHeadFor} (or \meth{searchModifierFor}) messages, first,
are reissued to the {\it textually} immediately preceding 
container actor which then forwards these messages to its {\it historically}
previous container actor.  This actor contains the head-centered
results of the analysis of the left context prior to the structural
extension held by the historical successor.\footnote{\label{headstuff}
Any container which holds the modifying part of the structural analysis of
the historical successor (in Fig.~\ref{back1}a this relates to ``the''
and ``silver'') is deleted. Hence, this deletion
renders the parser 
incomplete in spite of backtracking.  }
Attachments for heads or modifiers are now checked referring to
the historically preceding container and the active container as
depicted in Fig.~\ref{back1}a.

\begin{figure}
\centering
\mbox{
\epsfig{file=back1.eps,width=75mm}
\epsfig{file=back2.eps,width=90mm}}
\caption{Backtracking Mode
	\label{back2}\label{back1}}
\end{figure}

If the valency constraints are met, a new  phrase actor is formed
(cf.\ Fig.~\ref{back2}b) necessarily yielding a discontinuous analysis. 
A slightly modified protocol implements reanalysis, where the skipped
items send \meth{reSearchHeadFor} (or \meth{reSearchModifierFor})
messages to the new phrase actor, which forwards them directly to those word
actors where the discontinuity occurs.  As an example, consider the
fragment \bsp{the customer bought the silver} (with \bsp{silver}
in the noun reading, cf.\ Fig.~\ref{back1}a). This yields a
perfect analysis which, however, cannot be further augmented when the
word actor \bsp{notebook} asks for a possible
attachment.\footnote{Being
an
arc-eager parsing system, a \emph{possible} dependency relation will always be
established. Hence, the adjective reading of
\bsp{silver} will not be considered in the initial analysis.}  Two
intervening steps of reanalysis (cf.\ Fig.~\ref{back2}b
and~\ref{back2}c) yield the final structural
configuration depicted in Fig.~\ref{back2}d.

\def\lexwd{lexical item}


\paragraph{Prediction Protocol.}
The depth-first approach of the parser brings about a decision problem
whenever a phrase cannot be integrated into (one of) the left-context
analyses. 
Either, the left context and the current phrase are to be related by a word
not yet read from the input and, thus, the analysis should proceed without
an attachment.%
\footnote{
This effect occurs particularly often for verb-final languages such as
German. 
} 
Or, depth-first analysis was misguided and a backtrack should be invoked
to revise a former decision with respect to attachment information
available by now.

Prediction can be used to carry out a more informed selection between these
alternatives.  Words not yet read, but required for a complete analysis,
can be derived from the input analyzed so far, either top-down (predicting
a modifier) or bottom-up (predicting a head). Both types of prediction are
common in phrase-structure based parsers, e.g. Earley-style top-down
prediction \cite{earley70} or left-corner strategies with bottom-up
prediction \cite{kay80}. Since dependency grammars, in general, do not employ
non-lexical categories which can be predicted, so-called \emph{virtual
words} are constructed by the parser, which are later to be instantiated with
lexical content as it becomes available when the analysis proceeds.

\begin{figure}
\centering
\epsfig{file=predictN.eps,width=80mm}
\caption{Predicting and merging a noun
	\label{fig:np}}
\end{figure}

Whenever an active phrase cannot attach itself to the left context, the head
of this phrase may predict a virtual word  as
tentative head of a new phrase under which it is subordinated. The
virtual word is specified with respect to its word class,
morphosyntactic features, and order restrictions, but is left
vacuous with respect to its lexeme and semantic specification. In
this way, a determiner immediately constructs an NP (cf.\ Fig.\
\ref{fig:np}a), which can be attached to the left context and may
incrementally incorporate additional attributive adjectives until the
head noun is found (cf.\ Fig.\ \ref{fig:np}b).%
\footnote{ This procedure implements the notion of \emph{mother node
constructing categories} proposed by \textcite{hawkins94}, which are a
generalization of the notion \emph{head} to all words which
unambiguously determine their head. The linguistic puzzle about NP
vs.\ DP is thus solved. In contrast to Hawkins, we also allow for
multiple predictions.  }  The virtual word processes a
\meth{searchPredictionFor} protocol initiated by the next \lexwd. The
virtual word and this \lexwd\ are \emph{merged} iff the \lexwd\ is
at least as  specific as the virtual word (concerning word class
and features) and it is able to govern all
modifiers of the virtual word (cf.\ Fig.\ \ref{fig:np}c).

\begin{figure}
\centering
\epsfig{file=predictV.eps,width=100mm}
\caption{Predicting and splitting a verb
	\label{fig:vp}}
\end{figure}

This last criterion may not always be met, although the prediction, in
general, is correct. Consider the case of German verb-final
subclauses. A top-down prediction of the complementizer constructs a
virtual finite verb (designated by
\raisebox{-0.5ex}{\epsfig{file=blob2.eps,width=1em}}), which may
govern any number of NPs in the subclause (cf.\ Fig.\
\ref{fig:vp}a). If the verbal complex, however, consists of an infinite full
verb preceding a finite auxiliary, the modifiers of the virtual verb
must be distributed over two \lexwd s.%
\footnote{ We here assume  the finite auxiliary to govern the
subject (enforcing agreement), while the remaining complements are governed
by the infinite full verb.  } An extension of the prediction protocol
accounts for this case: A virtual word can be split if it may govern
the \lexwd\ and some modifiers can be transferred to the \lexwd. In
this case, the
\lexwd\  is subordinated to a newly created virtual
word (indicated by
\raisebox{-0.5ex}{\epsfig{file=blob3.eps,width=1em}} in
Fig.~\ref{fig:vp}b) governing the remaining modifiers. Since order
restrictions are available for virtual words, even non-projectivities
can be accounted for by this scheme (cf.\ Fig.\ \ref{fig:vp}b).%
\footnote{
Non-projectivities often arise, e.g. due to the fronting of a non-subject 
relative pronoun. As indicated by the dashed line in
Fig.~\ref{fig:vp}b and~\ref{fig:vp}c, we
employ additional projective relations to restrain ordering for
discontinuities. 
}

Although prediction allows parsing to proceed
incrementally and more informed (to the potential benefit of increased efficiency),
it engenders possible drawbacks: In underspecified contexts, a
lot of false predictions may arise and may dramatically increase
the number of ambiguous analyses. Furthermore, the introduction of
additional operations (prediction, split, and merge) increases the search
space of the parser. Part of the first problem is addressed by our
extensive usage of the word class hierarchy. If a set of predictions contains
all subclasses of some word class $W$, only one virtual word of class
$W$ is created.

\paragraph{Text Phenomena.}
A particularly interesting feature of the performance grammar we
propose is its capability to seamlessly integrate the sentence and
text level of linguistic analysis. We  have already alluded to the
notoriously intricate interactions between syntactic criteria and
semantic constraints at the phrasal and clausal level. The interaction
is even more necessary at the text level of analysis as semantic
interpretations have an immediate update effect on the discourse
representation structures to which text analysis procedures refer.
Their status and validity directly influence subsequent analyses at
the sentence level, e.g., by supplying proper referents for semantic
checks when establishing new dependency relations.  In addition,
lacking recognition and referential resolution of textual forms of
pronominal or nominal  anaphora \cite{strube.eacl95},
textual ellipses \cite{hahn.ecai96} and metonymies \cite{markert.ijcai97} leads to
invalid or incohesive text knowledge representation structures. These
not only yield invalid parsing results (at the methodological level)
but also preclude proper text knowledge acquisition (at the level of
system functionality).  Hence, we stress the neat integration of
syntactic and semantic checks during the parsing process at the
sentence and the text level.  We now turn to text grammar
specifications concerned with anaphora resolution and their
realization by a special text parsing protocol.

\begin{figure}
\centering
\fbox{\epsfig{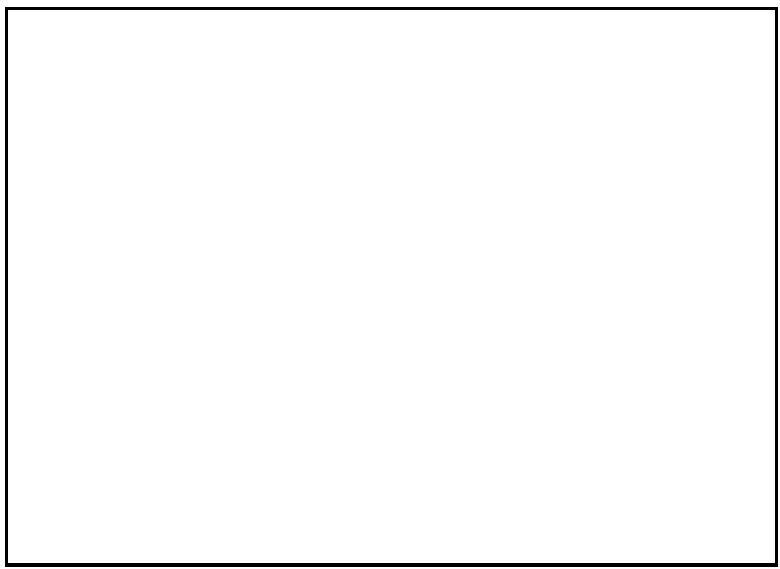}}
\caption{Anaphora Resolution Mode 
	\label{text}}
\end{figure}

\begin{figure}\centering
\fbox{\epsfig{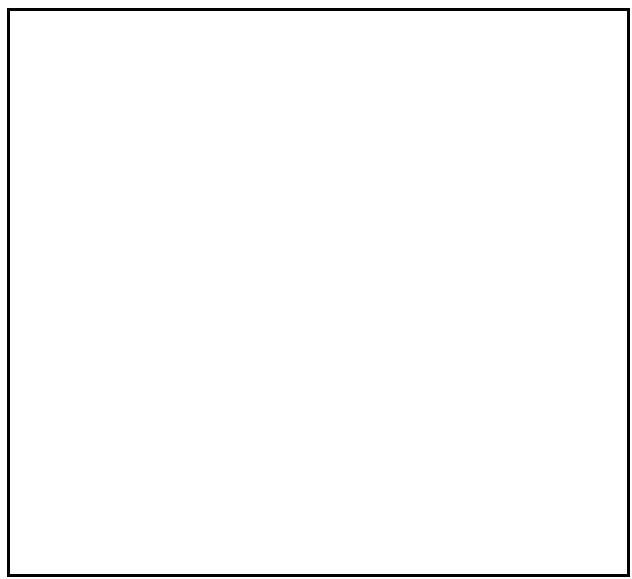}}
\caption{Sample Output of  Text Parsing\label{kb}}
\end{figure}

The protocol which accounts for local text coherence analysis makes
use of a special actor, the \emph{centering actor}, which keeps a
backward-looking center ($C_b$) and a preferentially ordered list of
forward-looking centers ($C_f$) of the previous utterance (we here
assume a functional approach \cite{strube.acl96} to 
 the well-known centering model originating from
\textcite{grosz_etal95}). These lists are accessed to establish proper
referential links between an anaphoric expression in the current
utterance and the valid antecedent in the preceding ones.  Nominal
anaphora (cf.\ the occurrences of \bsp{the company} and \bsp{these
printers} in Fig.\ \ref{text}) trigger a special
\meth{searchNomAntecedent} message. When it reaches the $C_f$ list,
possible antecedents are accessed in the given preference order.  If
an antecedent and the anaphor fulfill certain grammatical and
conceptual compatibility constraints,
an \meth{antecedentFound} message is issued to the anaphor, and finally,
the discourse referent of the antecedent replaces the one in the
original anaphoric expression in order to establish local coherence.
In case of successful anaphor
resolution an \meth{anaphorSucceed} message is sent from the resolved
anaphor to the centering actor in order to remove the determined
antecedent from the $C_f$ list (this avoids illegal follow-up
references). 
The effects of these changes at the
level of text knowledge structures are depicted in Fig.\ \ref{kb}, which
contains the terminological representation structures for the sentences in
Fig.\ \ref{text}.  



\section{Preliminary Evaluation}

Any text understanding system which is intended to meet the
requirements discussed in Section~\ref{ch:intro} faces severe
performance problems. Given a set of strong heuristics, a computationally
complete depth-first parsing strategy usually will increase the
parsing efficiency in the \emph{average case}, i.e., for input that is
in accordance with the parser's preferences. For the rest of the input
further processing is necessary.  Thus, the \emph{worst case} for a
depth-first strategy applies to input which cannot be assigned any
analysis at all (i.e., in cases of extra- or ungrammaticality). Such a
failure scenario leads to an exhaustive search of the parse
space. Unfortunately, under realistic conditions of real-world text input these
cases occur quite often. Hence, by using a computationally complete
depth-first strategy one merely would trade space complexity for time
complexity.

To maintain the potential for efficiency of depth-first operation it is
necessary to prevent the parser from exhaustive backtracking. In 
our approach this is achieved by two means.  First, by
restricting memoization of attachment candidates for backtracking
(e.g., by retaining only the head portion of a newly built phrase,
cf.\ footnote~\ref{headstuff}). Second, by restricting the
accessibility of attachment candidates for backtracking (e.g., by
bounding the forwarding of backtracking messages to linguistically
plausible barriers such as punctuation actors).  In effect, these
restrictions render the parser \emph{computationally incomplete},
since some input, though covered by the grammar specification, will not be
correctly analyzed. 


\subsection{Performance Aspects}

The stipulated efficiency gain that results from deciding against
completeness is empirically substantiated by a comparison of our
{\sc ParseTalk} system, henceforth designated as \emph{PT}, 
with a standard chart
parser,\footnote{
The active chart parser by \textcite{winograd83} was
adapted to parsing a dependency grammar. No packing or structure
sharing techniques could be used since the analyses have continuously
to be interpreted in conceptual terms.  We just remark that the
polynomial time complexity known from chart parsing of context-free
grammars does not carry over to linguistically adequate versions of dependency grammars \cite{neuhaus+broker97}.} abbreviated as
\emph{CP}.  As the CP does not employ any robustness mechanisms (one
might,e.g., incorporate those proposed by \textcite{mellish89}) the
current comparison had to be restricted to entirely grammatical
sentences. We also do not account for prediction mechanisms the
necessity of which we argued for in Section~\ref{grammar}. For the
time being, an evaluation of the prediction mechanisms is still under
way. Actually, the current comparison of the two parsers is based on a
set of 41 sentences from our corpus (articles from computer magazines)
that do not exhibit the type of structure requiring prediction (cf.\
Fig.~\ref{fig:vp} and the example therein).  For 40 of the test
sentences\footnote{The problem caused by the single missing sentence
is discussed in Section \ref{ling}.} the CP finds all correct analyses
but also those over-generated by the grammar. In combination, this
leads to a ratio of 2.3 of found analyses to correct ones. The PT
system (over-generating at a ratio of only 1.6) finds 36 correct
analyses, i.e., 90\% of the analyses covered by the grammar (cf.\ the
remark on 'near misses' in Section \ref{ling}). Our preliminary evaluation
study rests on two measurements, viz.\ one considering concrete
run-time data, the other comparing the number of method calls.

\begin{table}[h]\centering
\begin{tabular}{|c|c|r|}
\hline
number of & \multicolumn{2}{|c|}{speed-up factor}\\
samples & min--max & average\\\hline
25 & 1.1--4.2 & 2.8\\
10 & 5.1--8.9 & 6.9\\
6 & 10.9--54.8 & 45.2\\\hline
\end{tabular}
\caption{Ratio of run times of the CP and the PT system, chunked by
speed-up. \label{tab}} 
\end{table}

The loss in completeness is compensated by a reduction in processing
costs on the order of one magnitude on the average.  Since both systems
use the identical dependency grammar and knowledge representation the
implementation of which rests on identical Smalltalk and LOOM/Common
Lisp code, a run time comparison seems reasonable to some degree.  For
the test set the PT parser turned out to be about 17 times faster than
the CP parser (per sentence speed-up averaged at
over~10). Table~\ref{tab} gives an overview of the speed-up
distribution. 25 short to medium long sentences were processed with a
speed-up in a range from 1.1 to 4.2 times faster than the chart parser
averaging at 2.8. Another 10 longer and more complex sentences show
the effects of complexity reduction even  more
clearly, averaging at a speed-up of 6.9 (of
a range from 5.1 to 8.9). One of the remaining 6 very complex
sentences is discussed below.

Accordingly to these factors, the PT system spent
nearly two hours (on a SPARCstation~10 with 64 MB of main memory)
processing the entire test set, while the CP parser took more than 24
hours.  The exorbitant run times are largely a result of the
(incremental) conceptual interpretation, though these computations are
carried out by the LOOM system \cite{macgregor+bates87}, still one of
the fastest knowledge representation systems currently available
\cite{heinsohn_etal94}.

\begin{figure}
\centering
\epsfig{file=data.syn.eps,width=100mm}
\caption{Calls to \textsc{syntaxCheck}
	\label{data:syn}}
\end{figure}

\begin{figure}
\centering
\epsfig{file=data.con.eps,width=100mm}
\caption{Calls to \textsc{conceptCheck}
	\label{data:con}}
\end{figure}

While the chart parser is completely coded in Smalltalk, the PT system
is implemented in Actalk \cite{briot89} --- an extension of Smalltalk
which simulates the asynchronous communication and concurrent
execution of actors on sequential architectures. Thus, rather than
exploiting parallelism, the PT parser currently suffers from a
scheduling overhead. A more thorough comparison abstracting from these
implementational considerations can be made at the level of method
calls.  We here consider the computationally expensive methods
\textsc{syntaxCheck} and \textsc{conceptCheck} (cf.\
Section~\ref{grammar}). Especially the latter consumes large
computational resources, as mentioned above, since for each syntactic
interpretation variant a context has to be built in the KB system and
its conceptual consistency must be checked continuously. The number of
calls to these methods is given by the plots in Figs.~\ref{data:syn}
and \ref{data:con}. Sentences are ordered by increasing numbers of
calls to \textsc{syntaxCheck} as executed by the CP (this correlates
fairly well with the syntactic complexity of the input). The values for
sentences 39--41 in Fig.~\ref{data:syn} are left out in order to
preserve a proper scaling of the figure for plotting (39:
14389, 40=41: 27089 checks). A reduction of the total numbers of
syntactic as well as semantic checks by a factor of nine to ten can be
observed applying the strategies discussed for the PT system, i.e.,
the basic protocol plus skipping and backtracking.

\subsection{Linguistic Aspects}\label{ling}

The well-known PP attachment ambiguities pose a high processing burden
for any parsing system. At the same time, PP adjuncts often convey
crucial information from a conceptual point of view as in sentence 40:
\textsl{Bei einer Blockgr\"o\ss{}e, die kleiner als 32 KB ist,
erreicht die Quantum-Festplatte beim sequentiellen Lesen einen
Datendurchsatz von 1.100 KB/s bis 1.300 KB/s.} \textit{[For a
block size of less than 10 KB, the Quantum hard disk drive reaches a
data throughput of 1.100 KB/s to 1.300 KB/s for sequential
reading]}. Here, the chart parser considers all 16 globally ambiguous
analyses stemming from ambiguous PP attachments.

Apart from the speed-up discussed above the PT parser behaves
robust in the sense that it can gracefully handle cases of
underspecification or ungrammaticality. For instance, sentence 36
(\textsl{Im direkten Vergleich zur Seagate bietet sie f\"ur denselben
Preis weniger Kapazit\"at.} \textsl{[In direct comparison to
the Seagate drive, it (the tested drive) offers less capacity for the
same price.]})  contains an unspecified word 'weniger' (i.e. 'less')
such that no complete and correct analysis could be produced. Still,
the PT parser was able to find a 'near miss', i.e., a \emph{discontinuous}
analysis skipping just that word.

A case where the PT parser failed to find the correct analysis was
sentence 39: \textsl{Die Ger\"auschentwicklung der Festplatte ist
deutlich h\"oher als die Ger\"auschentwicklung der Maxtor 7080A.}
\textsl{[The drive's noise level is clearly higher than the noise
level of the Maxtor 7080A]}. When the adverb 'deutlich'
(i.e. 'clearly') is processed it is immediately attached to the matrix
verb as an adjunct. Actually it should modify 'h\"oher'
(i.e. 'higher'), but as it is not mandatory no backtrack is initiated
by the PT parser to find the correct analysis.

\section{Conclusion}

The incomplete depth-first nature of our approach leads to a
significant speed-up of processing approximately in the order of one
magnitude, which is gained at the risk of not
finding a correct analysis at all. This lack of completeness resulted
in the loss of about 10\% of the parses in our experiments and
correlates with fewer global ambiguities. We expect to 
find even more favorable results for the PT system 
when processing the complete corpus, i.e., when processing
material that requires prediction mechanisms. 


\medskip

\renewcommand{\baselinestretch}{0.9}
\noindent\small
{\bf Acknowledgments.} We would like to thank our
colleagues in the CLIF
group for fruitful discussions and instant support.
P.\ Neuhaus was supported by a grant from the interdisciplinary Graduate Program
{\em ``Menschliche und maschinelle Intelligenz''} 
({\em ``Human und machine intelligence''} at Freiburg University,
N.\ Br{\"o}ker was partially supported by a grants from DFG
(Ha 2097/1-3).


\end{document}